\documentclass{article} 
\usepackage{main,times}

\usepackage{amsmath,amsfonts,bm}

\def\eqref#1{equation~\ref{#1}}

\def\1{\bm{1}}

\DeclareMathAlphabet{\mathsfit}{\encodingdefault}{\sfdefault}{m}{sl}
\SetMathAlphabet{\mathsfit}{bold}{\encodingdefault}{\sfdefault}{bx}{n}

\definecolor{cvprblue}{rgb}{0.21,0.49,0.74}
\usepackage[pagebackref,breaklinks,colorlinks,citecolor=cvprblue]{hyperref}
\usepackage{url}
\usepackage{multirow}
\usepackage{multicol}
\usepackage{booktabs}
\usepackage{graphicx}
\usepackage{makecell}
\usepackage{marvosym}
\newcommand{\cN}{\mathcal{N}}
\newcommand{\I}{\boldsymbol{I}}
\newcommand{\0}{\mathbf{0}}

\title{FreeVS: Generative View Synthesis on Free Driving Trajectory}

\author{
Qitai Wang$^{1,2}$, Lue Fan$^{2,3}$, Yuqi Wang$^{1,2}$, Yuntao Chen$^{4}$\textsuperscript{\Letter}, Zhaoxiang Zhang$^{1,2,4}$\textsuperscript{\Letter} \\
$^1$School of Future Technology, University of Chinese Academy of Sciences (UCAS), \\
$^2$NLPR, MAIS, Institute of Automation, Chinese Academy of Sciences (CASIA),\\ 
$^3$CUHK
$^4$Center for Artificial Intelligence and Robotics, HKISI, CAS\\
\texttt{\{wangqitai2020, lue.fan, wangyuqi2020, zhaoxiang.zhang\}@ia.ac.cn,} \\
\texttt{chenyuntao08@gmail.com}\\
\small{Project Page: \url{https://freevs24.github.io/}} 
}

\newcommand{\change}[1]{\textcolor{black}{#1}}

\def\methodname{FreeVS }
\def\methodnamenospace{FreeVS}

\iclrfinalcopy 
\begin{document}

\maketitle

\begin{abstract}
Existing reconstruction-based novel view synthesis methods for driving scenes focus on synthesizing camera views along the recorded trajectory of the ego vehicle. 
Their image rendering performance will severely degrade on viewpoints falling out of the recorded trajectory, where camera rays are untrained.
We propose \methodnamenospace, a novel fully generative approach that can synthesize camera views on free new trajectories in real driving scenes. 
To control the generation results to be 3D consistent with the real scenes and accurate in viewpoint pose, we propose the pseudo-image representation of view priors to control the generation process.
Viewpoint transformation simulation is applied on pseudo-images to simulate camera movement in each direction.
Once trained, \methodname can be applied to any validation sequences without reconstruction process and synthesis views on novel trajectories.
Moreover, we propose two new challenging benchmarks tailored to driving scenes, which are novel camera synthesis and novel trajectory synthesis, emphasizing the freedom of viewpoints.
Given that no ground truth images are available on novel trajectories, we also propose to evaluate the consistency of images synthesized on novel trajectories with 3D perception models.
Experiments on the Waymo Open Dataset show that \methodname has a strong image synthesis performance on both the recorded trajectories and novel trajectories. 
\end{abstract}

\begin{figure*}[h]
  \includegraphics[width=\textwidth]{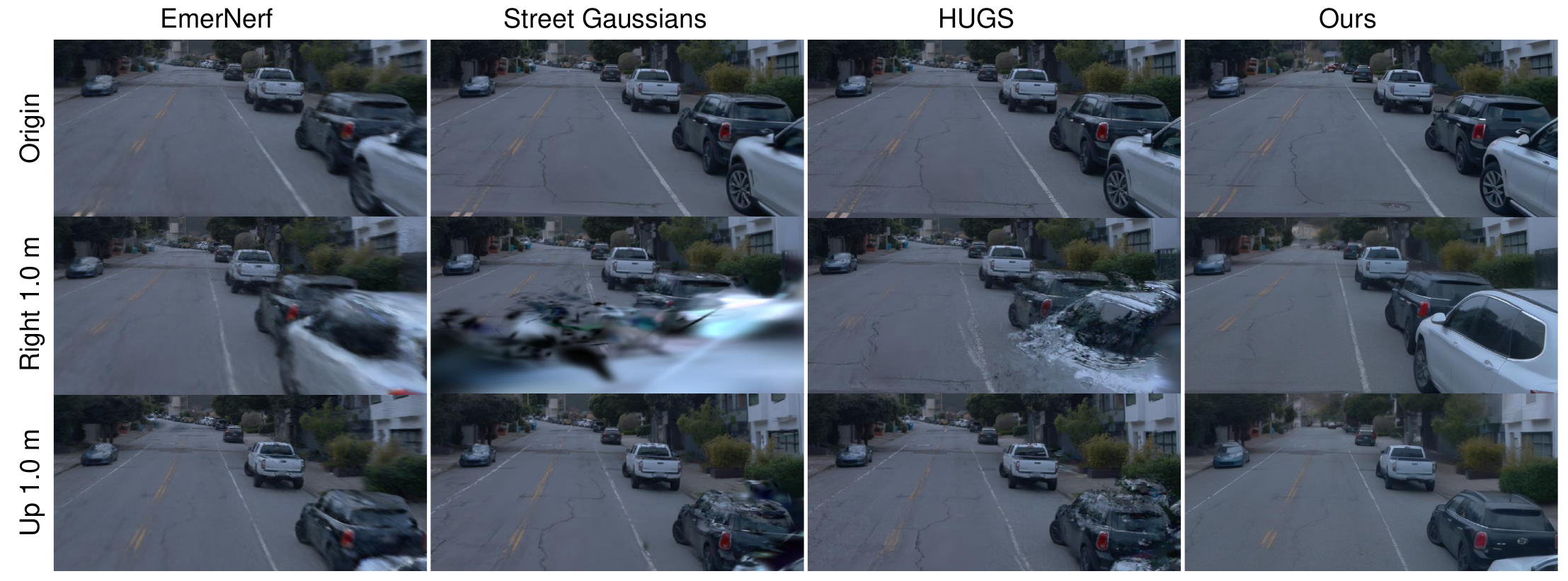}
  \caption{
  \textbf{Synthesis results comparison on the Waymo Open Dataset\citep{sun2020scalability}.} We show the camera views synthesized by NVS methods on the original front view (first row), viewpoint 1.0 m to the right (second row), and viewpoint 1.0 m above (third tow). Our method significantly outperforms previous NVS methods on viewpoint outside the existing ego trajectory.
  }
  \label{fig:teaser}
\end{figure*}

\section{Introduction}
\label{sec:intro}
Scene reconstruction and novel view synthesis (NVS) have gained special attention in embodied AI due to their potential to develop closed-loop simulations for embodied systems. 
Recent advances have led to remarkable improvements in the reconstruction quality of general scenes using multi-pass and multi-view recordings. 
However, reconstructing driving scenes presents distinct challenges due to the sparse observations inherent in their less controlled, real-world recording conditions.

Unlike general scene reconstruction settings, which typically leverage excessive views surrounding the scene, driving scene reconstruction generally only has access to image views along the single-pass ego driving trajectory. 
This limitation raises an important question: \emph{How well does driving scene reconstruction perform for novel viewpoints outside the recorded trajectory?}




Currently, existing driving scene NVS works\citep{guo2023streetsurf,wu2023mars,xie2023s,turki2023suds,yang2023emernerf,zhou2024drivinggaussian,chen2023periodic,yan2024street} only evaluate their image rendering quality along the recorded trajectory, leaving this question largely unanswered.
As shown in Fig.~\ref{fig:teaser}, the quality of rendering results of existing representative NVS methods degrades drastically when the rendering camera moves away from its recording trajectory. 
This is because, in driving scenes, recorded camera viewpoints are sparse in 3D space and homogeneous in their positions along the recorded trajectories. 
The sparsity and homogeneity of recorded camera views cause the camera rays shooting from the camera centers on novel trajectories largely untrained.

We propose \methodname to address this issue, which is a fully generative NVS method that can synthesize high-quality camera views inside and beyond the recorded driving trajectory.
We face two core challenges when building the \methodname.
The first challenge is accurately controlling the camera poses while maintaining the 3D geometry consistency of the generated views.
Although previous diffusion-based methods\citep{wang2023drivedreamer,lu2023wovogen,hu2023gaia,wang2024driving,yang2024generalized} are capable of controlling the camera motion in a coarse trajectory, their control precision is far from enough for safety-critical simulation purposes.
The second challenge is the ground truth images in the novel trajectories are unavailable, making it difficult to directly train a model to synthesize novel views beyond recorded trajectories.

To tackle the two challenges, the proposed \methodname leverages pseudo-image representation, a sparse yet accurate representation of 3D scenes obtained through colored 3D points projection.
Specifically, for each existing view, we create its pseudo-image counterpart by projecting colored point clouds into this view.
Here the colored points can be easily obtained by projecting point clouds to any valid images.
In this way, we obtain training data pairs to train a generative model that can generate a real image from its pseudo-image counterpart.
Since we create the pseudo images using ground truth camera models, they contain sparse but highly accurate appearance and geometry, sidestepping the tough challenge of accurately controlling the camera poses.
At inference time, we could create a pseudo-image for a novel viewpoint beyond the recorded trajectory and then synthesize the novel view using the trained generative model.
This design greatly narrowed the gap between synthesizing views inside and beyond the recorded trajectory.

To reveal the practicality of \methodnamenospace{}, we propose two challenging \change{benchmarks for evaluating the performance of NVS methods in driving scenes, which is more practically meaningful than the conventional evaluation on the recorded trajectories.}
\change{(i)} On the recorded trajectories, we propose the novel camera synthesis \change{benchmark}.
Instead of evaluating synthesis results on test frames sampled at intervals from video sequences (i.e. novel frame synthesis), we propose to drop \change{all images of a certain camera view (e.g. the front-side view) in the whole trajectory and synthesize the images of the dropped camera view.} 
\change{(ii)} We further propose the novel trajectory synthesis \change{benchmark}. 
With no ground truth images available on novel trajectories, we propose to evaluate the geometry consistency of synthesized views through the performance of off-the-shelf 3D detectors.
The experiments on the Waymo Open Dataset (WOD) demonstrate that \methodname outperforms previous NVS methods by a large margin in the \change{two more practical benchmarks} as well as in the traditional novel frame synthesis benchmark.


Our contributions are summarized as follows:
\begin{enumerate}
\item We propose \methodnamenospace{}, a fully generative view synthesis method for driving scenes that generate high-quality 3D-coherent novel views both for recorded and novel trajectories without time-cost reconstruction.
\item We devise two new benchmarks for evaluating driving NVS methods on novel trajectories beyond recorded ones.
\item Experiments on WOD show that \methodnamenospace{} achieves leading performance on synthesizing camera views inside and beyond the recorded trajectory.
\end{enumerate}

\section{Related Work}
\label{relatedworks}

\subsection{Novel View Synthesis through reconstruction} 
Recently, rapid progress has been achieved in novel view synthesis through 3D reconstruction and radiance field rendering.
Neural Radiance Fields (NeRF) ~\citep{mildenhall2020nerf} utilizes multi-layer perceptrons to represent continuous volumetric scenes and achieve a breakthrough in rendering quality.
Many works have extended NeRF to unbounded, dynamic urban scenes~\citep{tancik2022block,barron2022mip,ost2022pointlightfields,rematas2022urban,turki2022mega,lu2023dnmp,guo2023streetsurf,liu2023neural,wu2023mars,xie2023s,turki2023suds,yang2023unisim,wang2023neural,ost2021neural,tonderski2024neurad}.
Authors of MapNeRF~\citep{wu2023mapnerf} noticed the problem of NeRF in generating extrapolated views and proposed incorporating map priors to guide the training of radiance fields.
3D Gaussian Splatting~\citep{kerbl20233d} (3D GS) models scenes with numerous 3D Gaussians. Under this explicit representation, 3D GS can model scenes with significantly fewer parameters, while achieving faster rendering and training with splat-based rasterization.
3D GS is originally designed for static and bounded scenes.
Recently, some researchers have extended 3D GS to dynamic scenes~\citep{luiten2023dynamic,wu20234dgaussians,yang2023deformable3dgs,yang2023gs4d} and driving scenes~\citep{zhou2024drivinggaussian,chen2023periodic,yan2024street}. 
HUGS~\citep{zhou2024hugs} further jointly model the geometry, appearance, motion, and semantics in 3D scenes for better scene understanding.

\subsection{Novel View Synthesis through Generation}
Novel view synthesis through image generation has greatly benefitted from the advancements in diffusion models\citep{ho2020denoising,song2020denoising,rombach2022high,blattmann2023stable}. Zero-1-to-3\citep{liu2023zero} and ZeroNVS\citep{liu2023zero} generate novel views with a diffusion process conditioned on the reference image and the target camera pose embedded as a text embedding. GeNVS\citep{chan2023generative} condition the diffusion process on volume-rendered feature images. 
Reconfusion\citep{wu2024reconfusion} uses the diffusion model to refine images rendered by the reconstruction model as extra supervision to the reconstruction process.
Similarly, RealFusion\citep{melas2023realfusion} uses a diffusion model to provide an extra perspective view for object-centric reconstruction.
\cite{yu2024viewcrafter} use the Stable Video Diffusion model\citep{blattmann2023stable} to iteratively refine the rendered video along a novel camera trajectory based on the partial image wrapped from the reference view to the target view.
Most of the previous novel view synthesis through generation works are designed for object-centric\citep{liu2023zero,chan2023generative,wu2024reconfusion,melas2023realfusion,yu2024viewcrafter} or indoor\citep{liu2023zero,chan2023generative,wu2024reconfusion,yu2024viewcrafter} scenes.
For driving scenes, \cite{yu2024sgd} proposes SGD which generates novel views with a diffusion process conditioned on reference images and depth maps in the target view. However, SGD still only synthesizes camera views along the recorded trajectory of the ego vehicle.  Moreover, different from SGD which relies on the 3DGS model, \methodname is a fully generative model with performance comparable to reconstruction models even on recorded trajectories.

\section{\methodname}
\begin{figure*}[t]
  \includegraphics[width=\textwidth]{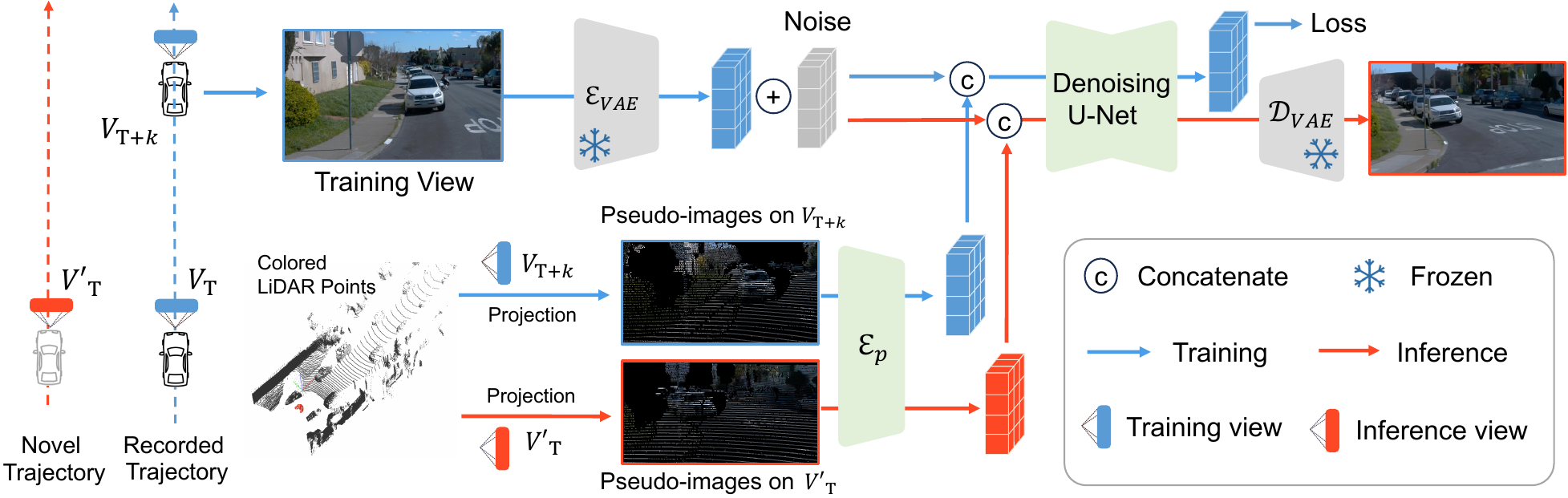}
  \vspace{-3mm}
  \caption{
  \textbf{Method pipeline of \methodnamenospace.} We propose to encode the view priors in driving scenes including appearance, 3D geometry, and pose of target viewpoints all in one modality as the pseudo-images. Best viewed in color. The diffusion model is trained to synthesize target views solely based on the unified pseudo-image priors. 
  }
  \label{fig:pipeline}
\end{figure*}

We introduce the detailed design of our proposed \methodname in this section.
We summarize the algorithm pipeline of \methodname in Fig.~\ref{fig:pipeline}.


\noindent\textbf{Overview of \methodnamenospace.} 
\methodname is a fully generative model that synthesizes new camera views on novel trajectories based on observations of 3D scenes from recorded trajectories.
\methodname is implemented as a conditional video diffusion model.
To ensure the model generates views from accurate viewpoints with consistent appearance attributes and 3D geometries as the real 3D scene, we formulate all essential priors regarding the 3D scene as pseudo-images to control the diffusion process.
Based on view prior conditions, \methodname is learned by denoising noised target views at training time and synthesizing target views from pure noise at inference time.

\subsection{View priors for view generation}
\paragraph{\change{Unified view prior representation.}} One major challenge of \change{generative} novel view synthesis is to ensure the generated \change{images} are consistent with the \change{priors in the novel view}.
\change{Here the view priors include the observed colors, 3D geometry, and camera pose of this view.
However, the different types of priors are in totally different modalities, posing a significant challenge for diffusion models to precisely encode them.
For example, as discussed in Sec.~\ref{sec:intro}, diffusion models cannot precisely control the camera motions (i.e., poses).
}
To tackle this challenge, we propose a pseudo-image representation that unifies all types of view priors in one modality.
Pseudo-images are obtained through colored point cloud projection.
Specifically, for each frame in a driving sequence,  we first merge LiDAR points across the nearby $r$ frames.
LiDAR points on moving objects will be merged along the moving trajectory of the object. 
Finally, we project the merged and colored LiDAR point cloud into the target camera viewpoints as pseudo-images. 
\change{In this way, we encode color information, geometry information, and the view pose into a unified pseudo-image, largely facilitating the learning of generative models. }

Compared with directly providing reference images and viewpoint transformations to the diffusion process, the pseudo-image representation greatly simplified the optimization objective of the generative model: With the former inputs, the model is required to have a correct understanding of the 3D scene geometry as well as the transformation of viewpoint to generate a correct view based on the reference image. 
In contrast, with pseudo-image as input, \methodname only needs to recover target views based on sparse valid pixels, which is more akin to a basic image completion task. 
The simplification of the training objective greatly enhances the model's robustness to unfamiliar viewpoint transformations, since the generated image is completed from sparse but geometrically accurate pixel points.

{\bf Viewpoint transformation simulation.} Another challenge of novel view generation on new trajectories stems from the absence of ground truth views beyond recorded trajectories. 
We can only train the generative model on recorded trajectories, where the diversity of viewpoint transformations is extremely limited.
For example, we have no access to the training sample where the frontal camera is moved laterally. 
However, such viewpoint transformation is essential for synthesizing views on novel trajectories at inference time. 
This brings a significant gap between training and inference for the generative model.
Moreover, we propose the viewpoint transformation simulation with pseudo-images.
\change{
At training time, we sample color and LiDAR priors from frames mismatched with the training image frames. That is to say, we force the generative model to recover current camera views based on observations from nearby frames.
}
Through this, we simulate the camera movement in each direction as a strong data augmentation on pseudo-image priors.
For example, as the ego vehicle moves along its heading direction, the side cameras are actually moving to their front or right,
Therefore although we have no access to the training data where the front camera is moved laterally, we can still simulate lateral camera movement by training \methodname on side-views with \change{mismatched observation-supervision frames}.

\subsection{Diffusion model for NVS}
\label{3.1}


{\bf Training of \methodname.} In each training iteration of \methodname, we randomly sample a colored LiDAR point cloud sequence ${\bf p} = (p_1,...p_n)$ from the driving scene dataset.
${\bf p}$ is a sequence of colored point clouds, each point cloud frame $p_i \in \mathbb{R}^{N_i \times 6}$ contains a set of 6-dimension 3D points.
3D points are recorded with their positions in the world reference frame and their visible colors.
From the driving sequence, we also sample a target camera viewpoint sequence ${\bf v} = ([v_1^1,...,v_1^m],...,[v_n^1,...,v_n^m])$ of $n$ frames and $m$ surrounding cameras. Each camera parameter $v_i$ stands for the intrinsics and extrinsics of a camera viewpoint. 
For a viewpoint in the target video with camera parameter $v_i^j$, we project the colored LiDAR point cloud $p_i$ into the viewpoint as pseudo-image $s_i^j = \text{Proj}(p_i,v_i^j)$. 
The training target of \methodname in each iteration is to recover the target images at sampled viewpoints based on the pseudo-image sequence ${\bf s} = ([s_1^1,...,s_1^m],...,[s_n^1,...,s_n^m]) \in \mathbb{R}^{n \times m \times 3 \times H\times W}$. 

During the training process of \methodname, the ground truth camera views ${\bf x} \in \mathbb{R}^{n \times m \times 3 \times H\times W}$ is also sampled along the viewpoint sequence ${\bf v}$.    
The ground truth camera views are encoded as target video latent representation $\mathcal{E}_{\text{VAE}}({\bf x}) = {\bf y} \in \mathbb{R}^{n \times m \times c \times h\times w}$ through an frozen VAE encoder. 
Then We have the diffused inputs ${\bf y}_r = \alpha_\gamma{\bf y}+\sigma_\gamma{\bf \epsilon},{\bf \epsilon} \sim \mathcal{N}({\bf 0}, \boldsymbol{I})$, here $\alpha_\tau$ and $\sigma_\tau$ is noise schedule at the diffusion time step $\tau$.
We also encode the pseudo-images into the latent representation $\mathcal{E}_p({\bf s}) = {\bf z} \in \mathbb{R}^{n \times m \times c \times h\times w}$ with a 2D encoder trained simultaneously with the diffusion model. 
We concatenate ${\bf y}_r$ and ${\bf z}$ as the input ${\bf k} \in \mathbb{R}^{n \times m \times 2c \times h\times w}$ to the diffusion model to predict the noise upon ${\bf y}$. 
We have a denoising model ${\bf f}_{\theta}$ with parameters $\theta$ that take ${\bf y}_r$,${\bf z}$ as inputs and optimized by minimizing the following denoising objective:
\begin{equation}
    \mathbb{E}_{\mathbf{k},\tau\sim p_\tau,\epsilon\sim\cN(\0, \I)}[\|\epsilon - {\bf f}_{\theta}({\bf k};{\bf c},\tau)\|_2^2],
    \label{eq:overall_formulation}
\end{equation}
Where ${\bf c}$ is the description conditions generated by encoding the reference camera views with an off-the-shelf CLIP-vision model\citep{radford2021learning}, following the convention of diffusion models. $p_\tau$ is a uniform distribution over the diffusion time $\tau$. 


{\bf Synthesizing views on novel trajectories with \methodname.} During the inference process of \methodname, we project the colored LiDAR points in each frame into the targeted camera poses to generate pseudo-image sequence for image synthesis.
The diffusion model is fed with pure noise latents concatenated with pseudo-image latents. 
The diffused latent will be decoded as synthesized views through an off-the-shelf VAE decoder $\mathcal{D}_{\text{VAE}}$.



\subsection{Evaluating NVS on novel camera and novel trajectory synthesis} 
\begin{figure*}[t]
\centering
  \includegraphics[width=0.9\textwidth]{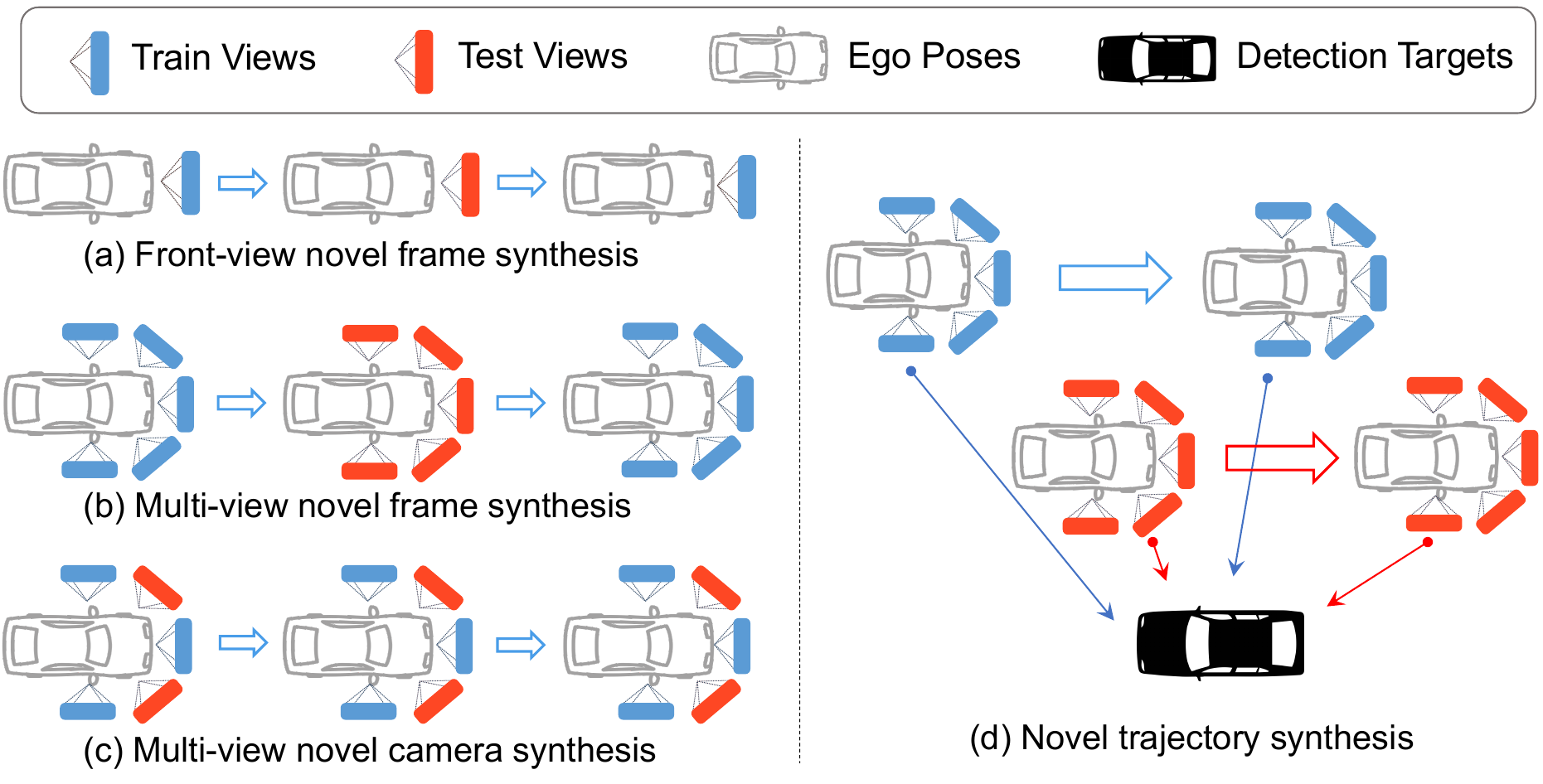}
  \vspace{-3mm}
  \caption{
  \textbf{Benchmarks for evaluating NVS methods in driving scenes.} We conclude the previous NVS evaluation benchmarks for driving scenes as (a) and (b). We propose two novel evaluation benchmarks: the novel camera synthesis benchmark as in (c), and the novel trajectory synthesis benchmark as in (d). Best viewed in color.
  }
  \vspace{-4mm}
  \label{fig:evalsetting}
\end{figure*}

To thoroughly demonstrate the view generalization capability of our \methodname, which can truly meet the demands of closed-loop embodied simulation, we present a comprehensive discussion of evaluation benchmarks for novel view synthesis in driving scenes. Fig. \ref{fig:evalsetting} illustrates this: panels (a) and (b) summarize existing evaluation benchmarks, while panels (c) and (d) introduce our two new challenging NVS benchmarks.

{\bf Evaluating NVS on recorded trajectories.} All current NVS works for driving scenes evaluate their NVS performance on test frames sampled periodically along the recorded trajectory.
Some previous driving scene NVS methods, such as Street Gaussians\citep{yan2024street}, NSG\citep{ost2021neural}, and Mars\citep{wu2023mars}, evaluate their performance with only front camera views considered, as illustrated in Fig.~\ref{fig:evalsetting}(a).
Other NVS methods take the multi-view cameras into consideration as illustrated in Fig.~\ref{fig:evalsetting}(b), such as DrivinGaussian\citep{zhou2024drivinggaussian}, 
PVG\citep{chen2023periodic}, EmerNerf\citep{yang2023emernerf}, NeuRAD\citep{tonderski2024neurad}, S-Nerf\citep{xie2023s}, and SUDS\citep{turki2023suds}.
All these two evaluation benchmarks sample test frames periodically along the trajectory, i.e. {\bf novel frame synthesis}.
In such cases, camera views in test frames can be directly inferred from the adjacent frames, especially for datasets with a high video frame rate (e.g. 10Hz for the WOD dataset). 
To provide a more challenging evaluation setting for driving scene NVS methods, we propose the {\bf novel camera synthesis benchmark} as illustrated in Fig.~\ref{fig:evalsetting}(c). 
Instead of periodically sampling test frames, we drop images collected by certain multi-view cameras throughout a driving sequence as test views. 
For example, for a driving sequence in the WOD dataset, we provide NVS methods with the front, and side camera views as training views and evaluate the synthesis results on front-left and front-right views.
Under the novel camera synthesis benchmark, NVS methods are required to synthesize views on unseen camera poses, which places higher demands on accurately modeling the 3D scene. 
We ensure in the validation sequences, most 3D contents in front-side cameras are observed in the front or side camera views along the ego trajectory.

{\bf Novel trajectory synthesis.} 
On test views sampled from the recorded trajectories, the ground truth camera images are available for evaluating the synthesized images with image similarity metrics including SSIM, PSNR, and LPIPS\citep{zhang2018unreasonable}.
Differently, in driving scenes, there are no ground truth images available on novel trajectories.
The Fréchet Inception Distance (FID)\citep{Seitzer2020FID} metric can compare the overall image distribution between synthesized images on novel trajectories and ground truth images on recorded trajectories, but it can not assess the fidelity of the synthesized images to the 3D scenes at all.
In addition to qualitative visualization comparisons, we also propose the {\bf perceptual robustness evaluation} to assess the geometry consistency performance of NVS methods on new ego trajectories.

In driving scenes, modern image-based 3D perception models have achieved high robustness. 
As shown in Fig.~\ref{fig:evalsetting}(d), assuming an NVS method can synthesize views on a novel trajectory with ideal image quality, the perception model feed with synthesized views should still be able to produce accurate predictions on the novel trajectory.
With such an assumption, we believe that the performance of an off-the-shelf perception model on novel trajectories can reflect the quality of images synthesized by the NVS methods.
Under the novel trajectory synthesis benchmark, we feed the synthesized images and camera poses on the novel trajectory to an off-the-shelf 3D camera-based detector. 
The detection results are evaluated with the longitudinal error tolerant mean average precision (LET-mAP)\citep{hung2022let} metric on the WOD dataset. 
For all NVS methods, we modify novel trajectories by laterally shifting the ego positions in each frames. We shift the trajectories by 1.0 m, 2.0 m, and 4.0 m and report the mean evaluated results as $\text{mAP}^{\text{LET}}_{1.0m}$,$\text{mAP}^{\text{LET}}_{2.0m}$, and $\text{mAP}^{\text{LET}}_{4.0m}$.

\section{Experiments}
In this section, we first introduce our experimental setup including datasets, evaluation benchmarks, method implementation details, and counterpart methods. 
Then we provide the quantitative and qualitative experiment results. 

\noindent\textbf{Datasets.}
We perform experiments on the WOD dataset\citep{sun2020scalability}. We select 12 driving sequences for evaluating NVS methods. We ensure that there is ample space on both sides of the ego vehicle in most frames of the selected sequence to simulate novel trajectories by lateral moving the ego vehicle. For each sequence, all 200 data frames in 10Hz are used. 

\noindent\textbf{Evaluation of NVS methods.}
We compare \methodname with NVS counterparts under all the experiment benchmarks shown in Fig.~\ref{fig:evalsetting}. 
For the front-view or multi-view novel frame synthesis benchmark (Fig.~\ref{fig:evalsetting}(a) and (b)), we sample every fourth frame in driving sequences as test frames. 
All the remaining frames are used for training NVS counterparts, or as input frames for \methodnamenospace.
On, we report metrics including SSIM, PSNR, and LPIPS. 
Under the novel camera synthesis benchmark, we reserve all the front-side camera views as test views and use the front and side camera views as train views throughout each sequence.
Note that for \methodname which does not require scene reconstruction on validation sequences, we only take information from the train views to generate test views.

Moreover, we also evaluate NVS methods on novel trajectories with the FID score and the proposed perceptual robustness evaluation method.
We take MV-FCOS3D++\citep{wang2022mvfcos3d++}, a basic yet strong multi-view camera-based 3D detector as our baseline detection model. We follow most of the settings of the official open-sourced implementation of MV-FCOS3D++.
We train MV-FCOS3D++ for 24 epochs on the WOD training set (except for the validation sequences in our experiments) to obtain the baseline detector.
Following \citep{wang2022mvfcos3d++}, we initialize MV-FCOS3D++ from an FCOS3D++ checkpoint, which is also trained on the above training sequences.
We feed the camera views synthesized on novel trajectories to the baseline detector. 
We report camera-based 3D detection metrics LET-mAP\citep{hung2022let} on the vehicle class as $\text{mAP}^{\text{LET}}$. 

\noindent\textbf{Method details.}
We implement the proposed \methodname pipeline based on Stable Video Diffusion (SVD)\citep{blattmann2023stable}. 
We initialize the diffusion model from a pre-trained Stable Diffusion checkpoint\citep{rombach2022high}. 
\methodname is trained on the WOD training set, except for the selected validation sequences.
To generate pseudo-images, we accumulate colored LiDAR points across the adjacent $\pm2$ frames of each frame. 
If a 3D LiDAR point has more than one projected 2D point in multi-view images, the mean color of its projected image points will be recorded. 
For viewpoint transformation simulation, we randomly sample the target viewpoint sequence starting from the adjacent $\pm4$ frames of the first frame of the source point cloud sequence. We employ a ConVNext-T\citep{liu2022convnet} backbone as the pseudo-image encoder. We train the model for 40,000 iterations with a batch size of 8 and video frame length $n=8$. Please refer to the appendix for more training details.

\noindent\textbf{Novel view synthesis counterparts.} 
We compare our novel view synthesis method with the 3DGS\citep{kerbl20233d}, EmerNeRF\citep{yang2023emernerf}, and Street Gaussians\citep{yan2024street}. 
All counterpart methods are implemented based on their official implementation. 
For 3DGS which is not initially designed for unbounded driving scenes, we largely increase its max training iterations for better convergence of the model. Please check the appendix for more implementation details on NVS counterparts.

\subsection{SOTA comparison under the proposed challenging new benchmarks.}
{\bf Novel camera synthesis.} We first report the performance of NVS methods under the proposed multi-view novel camera synthesis benchmark in Table \ref{tab:novelcamera}.
\methodname achieves leading performance on all metrics by a large margin.
Previous NVS methods tend to render images with severe image distortion or massive unnatural artifacts when facing severe scene information loss on the target views, as shown in Fig.~\ref{fig:visualize1}.
Meanwhile, \methodname can generate camera views close to ground truth views based on limited 3D scene observations.

{\bf Novel trajectory synthesis.} We also report the FID and perceptual robustness performance of NVS methods on novel trajectories in Table \ref{tab:main_newtraj}. 
The proposed \methodname outperforms previous NVS methods on almost all metrics with different lateral offsets applied to the viewpoints. 
Compared to previous NVS methods, the proposed \methodname has a very strong performance on the FID metric. 
This is mainly because the proposed \methodname is nearly free from image degradation and artifacts when synthesizing images on novel trajectories. 
\methodname also has the strongest $\text{mAP}^{\text{LET}}$ performance among all NVS methods, which indicates that as a generation-based method, \methodname is of even higher fidelity to the 3D scene geometry compared with previous reconstruction-based methods when rendering views on novel trajectories. We also provide a visualization comparison example in Fig.~\ref{fig:visualize2}.

While \methodname relies on LiDAR point inputs, EmerNerf and Street Gaussians also rely on LiDAR depth supervision during their training process.
Therefore \methodname did not gain any information advantages in our experiments.
Moreover, as a fully generative method, \methodname does not require any scene reconstruction process when applied to validation sequences.
From this aspect, at inference time, \methodname costs less computational resources even compared with 3DGS-based methods, which usually take 1-2 hours to model a validation sequence of 20s.

\begin{table*}[t]\footnotesize
\centering
\caption{
{\bf Comparison with NVS counterparts on novel camera synthesis.} For all NVS methods, we use all front and side camera views as source views to synthesize the front-side camera views.}
\setlength{\tabcolsep}{6pt}
\begin{tabular}{c|ccc}
\toprule
   & \multicolumn{3}{c|}{Front-side camera syntheising} \\ 
   \multirow{-2}{*}{Methods} & SSIM$\uparrow$ & PSNR$\uparrow$ & LPIPS$\downarrow$ \\
\midrule
   3D-GS\citep{kerbl20233d} & 0.484 & 15.97 & 0.442 \\%
   EmerNerf\citep{yang2023emernerf} & 0.603 & 19.61 & 0.330 \\
   StreetGaussian\citep{yan2024street} & 0.531 & 17.35 & 0.397\\
   Ours & {\bf 0.628} & {\bf 20.70} & {\bf 0.283}\\

\bottomrule
\end{tabular}
\label{tab:novelcamera}
\end{table*}

\begin{table*}[t]\footnotesize
\centering
\caption{
{\bf Comparison with NVS counterparts on novel trajectories.} The $y$ axis is defined lateral to the ego vehicle's heading direction. \dag: performance of baseline detector on ground truth images. 
 }
\setlength{\tabcolsep}{6pt}
\resizebox{\linewidth}{!}{
\begin{tabular}{c|cc|cc|cc|cc}
\toprule
   & \multicolumn{2}{c|}{$y\pm0.0 m$} & \multicolumn{2}{c|}{$y\pm1.0 m$} & \multicolumn{2}{c|}{$y\pm2.0 m$} & \multicolumn{2}{c}{$y\pm4.0 m$} \\ 
   \multirow{-2}{*}{Methods} & FID$\downarrow$ & $\text{mAP}^{\text{LET}}$$\uparrow$ & FID$\downarrow$ & $\text{mAP}^{\text{LET}}_{1.0m}$$\uparrow$ & FID$\downarrow$ & $\text{mAP}^{\text{LET}}_{2.0m}$$\uparrow$ & FID$\downarrow$ & $\text{mAP}^{\text{LET}}_{4.0m}$$\uparrow$\\
\midrule
    GT\dag & - & 0.895 & - & - & -& -& -& -\\
\midrule
   3D-GS & 34.79 & 0.729 & 52.07 & 0.605 & 61.16 & 0.581 & 86.21 & 0.452 \\
   EmerNerf & 53.88 & 0.600 & 58.26 & 0.510 & 69.50 & 0.478 & 84.81 & 0.464 \\
   StreetGaussian & 21.62 & {\bf 0.826} & 41.17 & 0.738 & 55.71 & 0.682 & 80.44 & 0.544 \\
   Ours & {\bf 11.18} & 0.816 & {\bf 13.45} & {\bf 0.786} & {\bf 16.60} & {\bf 0.724} & {\bf 22.08} & {\bf 0.612} \\
\bottomrule
\end{tabular}
} 
\label{tab:main_newtraj}
\end{table*}
\begin{table*}[t]\footnotesize
\centering
\caption{
{\bf Comparison with NVS counterparts on recorded trajectories.}
We report the performance of NVS methods when only the front-view cameras are considered or when all multi-view cameras are considered. Reconstruction time cost and FPS are measured under the multi-view setting, with a single NVIDIA L20 GPU. \dag: we largely increased the max training iterations of 3D-GS for better performance in driving scenes.
}
\setlength{\tabcolsep}{4pt}
\resizebox{\linewidth}{!}{
\begin{tabular}{c|ccc|ccc|cc}
\toprule
   & \multicolumn{3}{c|}{Front View} & \multicolumn{3}{c|}{Multi-view} \\ 
   \multirow{-2}{*}{Methods} & SSIM$\uparrow$ & PSNR$\uparrow$ & LPIPS$\downarrow$ & SSIM$\uparrow$ & PSNR$\uparrow$ & LPIPS$\downarrow$ & \multirow{-2}{*}{\makecell[c]{Reconstruction\\
   time}} & \multirow{-2}{*}{FPS}\\
\midrule
   3D-GS\citep{kerbl20233d} & 0.799 & 26.31 & 0.143 & 0.586 & 19.21 & 0.366 & 2-3h\dag & \textbf{61.2}\\
   EmerNerf\citep{yang2023emernerf} & 0.869 & 30.28 & 0.155 & 0.689 & 24.68 & 0.347 & 2-3h & 0.2 \\
   StreetGaussian\citep{yan2024street} & {\bf 0.903} & {\bf 30.80} & {\bf 0.096} & 0.702 & 22.47 & 0.314 & 1-2h & 52.6\\
   Ours & 0.787 & 25.30 & 0.139 & {\bf 0.730} & {\bf 24.96} & {\bf 0.179} & - & 0.9\\

\bottomrule
\end{tabular}
} 
\label{tab:mainpsnr}
\end{table*}
\subsection{SOTA comparison on novel frame synthesis}
We also report the performance of NVS methods under the traditional front-view novel frame synthesis or multi-view novel frame synthesis benchmark in Table \ref{tab:mainpsnr}. 
The performance of previous NVS methods is strong when only the front-view camera is considered. 
However, when it comes to the multi-view setting which is more aligned with the current autonomous driving scenes, the performance of previous NVS methods is surpassed by the proposed \methodname by a large margin.
It is worth mentioning that in Table \ref{tab:mainpsnr}, all previous reconstruction-based NVS methods exhibit a significant performance drop when multi-view cameras are considered. 
We think this is due to the increased number of training views, the expanded range of the visible 3D scene, and the rapidly changing content in lateral views, all of which make the convergence of reconstruction models more difficult.

\begin{table*}[t]\footnotesize
\centering
\caption{
{\bf Ablation study on view prior condition.} We conduct a breakdown ablation on the proposed pseudo-image representation of view priors. The $y$ axis is defined lateral to the ego vehicle's heading direction.
 } 
\setlength{\tabcolsep}{6pt}
\begin{tabular}{c|c|c|ccc|cc}
\toprule
      & \multirow{2}{*}{View priors} & \multirow{2}{*}{Encoders} &  \multicolumn{3}{c|}{Multi-view} & \multicolumn{2}{c}{$y\pm2.0m$}\\ 
      & &  & SSIM$\uparrow$ & PSNR$\uparrow$ & LPIPS$\downarrow$ & FID$\downarrow$ & $\text{mAP}^{\text{LET}}_{2.0m}$$\uparrow$ \\
\midrule
   (a) & full priors & 2D-Conv & 0.704 & 23.28 & 0.203 & 21.27 & 0.690 \\  
   (b) & w/o. color & 2D-Conv & 0.701 & 23.05 & 0.231 & 23.13 & 0.687\\ 
   (c) & w/o. LiDAR & 2D-Conv + MLP & 0.613 & 19.86 & 0.288 & 21.25 & 0.013\\ 
   (d) & w/o. projection & 2D-Conv + 3D + MLP & 0.609 & 19.88 & 0.284 & 21.32 & 0.028\\
   (e) & full priors & 2D-Attn & 0.706 & 23.27 & 0.202 & 21.25 & 0.692\\  
\bottomrule
\end{tabular}
\label{tab:abl_3dpripr}
\end{table*}
\begin{table*}[h]\footnotesize
\centering
\caption{
{\bf Ablation study on viewpoint translation simulation.} The $y$ axis is defined lateral to the ego vehicle's heading direction.
 }
\setlength{\tabcolsep}{6pt}
\begin{tabular}{c|ccc|cc}
\toprule
    &  \multicolumn{3}{c|}{Multi-view} & \multicolumn{2}{c}{$y\pm2.0m$}\\ 
   \multirow{-2}{*}{Temporal sampling window size} & SSIM$\uparrow$ & PSNR$\uparrow$ & LPIPS$\downarrow$ & FID$\downarrow$ & $\text{mAP}^{\text{LET}}_{2.0m}$$\uparrow$ \\
\midrule
   -  & 0.733 & 25.04 & 0.180 & 18.29 & 0.704  \\
   $\pm2$ frames & 0.734 & 25.04 & 0.179 & 18.24 & 0.710 \\
   $\pm4$ frames &  0.730 & 24.96 & 0.179 & 18.01 & 0.722\\
   $\pm8$ frames & 0.717 & 24.82 & 0.188 & 18.09 & 0.720 \\
\bottomrule
\end{tabular}
\label{tab:abl_tempotalaug}
\end{table*}
\subsection{Ablation Studies}
\label{4.3}
\noindent\textbf{Ablation on view prior condition.}
We first ablate on the representation of view prior as conditions for the diffusion process, as shown in Table \ref{tab:abl_3dpripr}.
We apply a breakdown experiment on the pseudo-image representation.
Models are trained for 20,000 iterations.
We start by dropping the color information in the pseudo-image representation, represented by Table \ref{tab:abl_3dpripr}(b).
Dropping the color nearly does not affect the geometric accuracy of rendered results, but has a considerable impact on the image similarity metrics.
Then we experiment with dropping the LiDAR inputs (c), where the reference images and camera pose transformation matrix (from the reference view to the target view) are independently encoded by a VAE or MLP encoder. Under this setting, we found the diffusion model unable to accurately synthesize views on the target viewpoint. Most time, the model ignores the pose condition and moves the camera viewpoint by its familiar viewpoint transformation. (e.g. always move the frontal camera forward or backward, or move the side camera left or right.)  Based on (c), we experiment with preserving all view prior inputs but do not unify them as pseudo-images (d). The LiDAR points are encoded as latents with a point cloud backbone\citep{yan2018second}. The experiment result shows the model fails at utilizing LiDAR inputs due to its significant gap with 2D images, the model trained under setting (d) has an identical performance as the mode trained under setting (c). Due to the wrong viewpoint of most generated images, trained models under settings (c) and (d) have extremely poor perceptual robustness performance. Through this observation, we can conclude that the pseudo-image representation greatly improved the overall quality and viewpoint controllability of images generation. 
Finally, we compare encoding the pseudo-images with a 2D-Conv encoder or an 2D-attention encoder\citep{liu2021swin} with setting (e). 




\noindent\textbf{Ablation study on viewpoint transformation simulation.} 
Viewpoint transformation simulation aims at constructing source and target frame pairs from recorded sequences to simulate camera movement in various directions. 
We report the results of ablation studies on viewpoint transformation simulation with pseudo-images in Table \ref{tab:abl_tempotalaug}. 
We report the performance of \methodname under the multi-view novel frame synthesis setting and on novel trajectories generated by applying  2.0 m lateral offsets to the recorded trajectories.
As shown in Table \ref{tab:abl_tempotalaug}, sampling target frames from adjacent $\pm2$ or $\pm4$ frames from the source frame can boost the view-synthesize performance of \methodname on novel trajectories.
When the temporal sampling window size exceeds $\pm4$ frames, the view-synthesize performance of \methodname on the recorded trajectory will be negatively affected. We believe this is due to the large timestamp mismatch between view priors and target images hindering the model's convergence. 



\vspace{-2mm}
\subsection{Visualization comparison}
\begin{figure*}[h]
  \includegraphics[width=\textwidth]{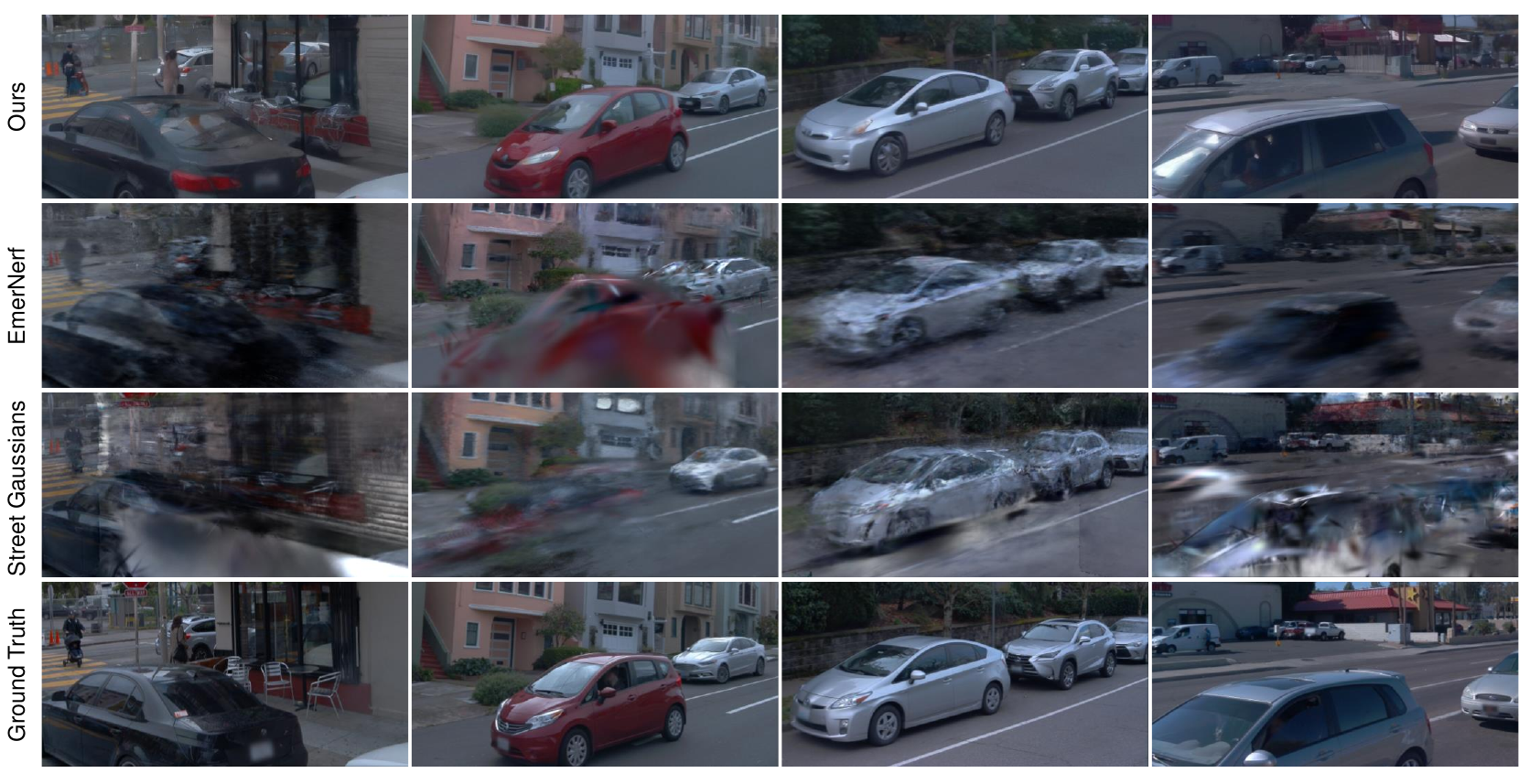}
  \vspace{-4mm}
  \caption{
  \textbf{Visualization comparison on novel-camera synthesis benchmark.} We show the front-side camera views synthesized from front and side camera views with NVS methods. 
  }
  \label{fig:visualize1}
\end{figure*}

\begin{figure*}[ht]
  \includegraphics[width=\textwidth]{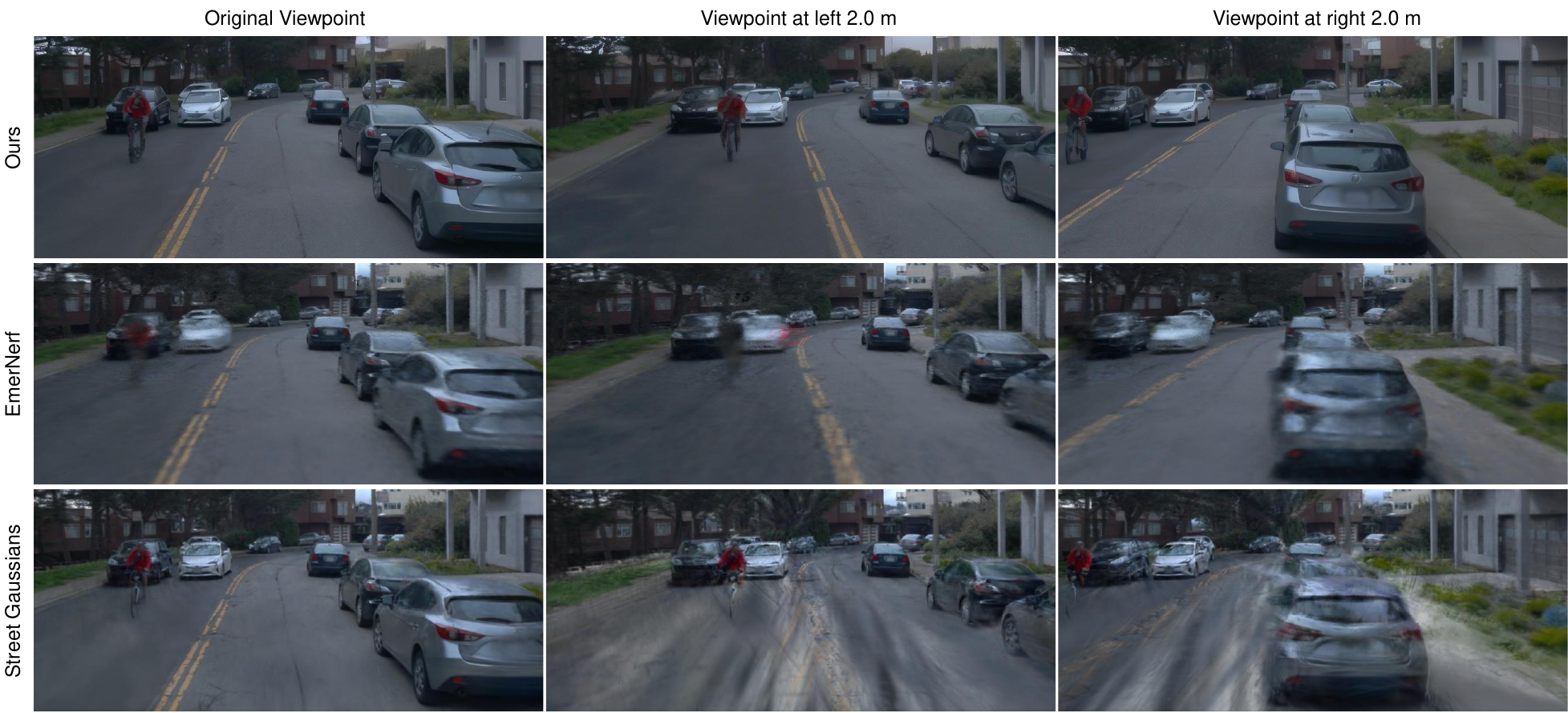}
  \vspace{-4mm}
  \caption{
  \textbf{Visualization comparison on novel trajectories.} We show the camera views synthesized by NVS methods on the original training viewpoint, viewpoint 2.0 m left of the original viewpoint, and viewpoint 2.0 m right of the original viewpoint.
  }
  \label{fig:visualize2}
\end{figure*}

We show visualization comparisons between NVS methods under the novel camera synthesis benchmark in Fig.~\ref{fig:visualize1}, and under the novel trajectory synthesis benchmark in Fig.~\ref{fig:visualize2}.

\vspace{-2mm}


\subsection{Conclusion}
We present \methodnamenospace, a novel fully generative method for synthesizing camera views on free driving trajectory.
We propose the pseudo-image representation of view priors, which conveys accurate 3D scene geometry and viewpoint conditions through colored point projection.
The diffusion model is trained to synthesize target views solely based on pseudo-images.
In this paper, we fully discuss the evaluation benchmarks for driving scene  NVS. 
We propose two novel evaluation benchmarks including the novel camera synthesis benchmark and the novel trajectory synthesis benchmark.
We also propose the perceptual robustness evaluation method for assessing the performance of NVS methods on novel trajectories.
Experiments across several experiment benchmarks show that \methodname achieves leading performance in synthesizing camera views inside or beyond recorded trajectories. 

\clearpage
\bibliography{main}
\bibliographystyle{main}

\clearpage
\appendix
\section{Appendix}
\setcounter{figure}{0}
\setcounter{table}{0}
\renewcommand\thefigure{\Alph{figure}}
\renewcommand\thetable{\Alph{table}}
\subsection{Implementation Details}
{\bf Implementation details of \methodnamenospace.}
We employ a ConVNext-T\citep{liu2022convnet} encoder to encode pseudo-images. 
For training \methodnamenospace, the diffusion model is initialized with Stable Diffusion checkpoints~\citep{rombach2022high}.
We train the model for 40,000 iterations with a batch size of 8 and video frame length $n=8$. We use the AdamW optimizer~\citep{kingma2014adam} with a learning rate $1\times10^{-4}$.
During training time, we randomly drop the pseudo-image condition latent as well as the CLIP text description latent with a probability of 20\%.
We enable the viewpoint transformation simulation with a probability of 50\%.
During inference, we set the number of sampling steps as 25 and stochasticity $\eta$=1.0.
When synthesizing images on the existing trajectory, we set the classifier-free guidance (CFG)\citep{ho2022classifier} guidance scale to 1.0. 
For synthesizing images on novel cameras and new trajectories, we enlarge the CFG guidance scale to 2.0 to strengthen the control of 3D prior conditions over the generated results.

{\bf Implementation details of NVS counterparts.}
For 3DGS which is not initially designed for unbounded driving scenes, we optimize its performance by adjusting its hyperparameters, including setting the densification interval to 500 iterations, setting the opacity reset interval to 10000, and training the 3DGS models for 100000 iterations while densifying 3D Gaussians until 50000 iterations.
We also noticed that the Street Gaussians models have a convergence issue when training with all 200 frames of each sequence.
Therefore we split the validation sequences into two 100-frame sequences for training the Street Gaussians models, following its official configuration\citep{yan2024street}.
As 3DGS and EmerNerf suffer from high memory costs when training with high-resolution images, we resize the input images on WOD from a resolution of $1920\times1280$ to $1248\times832$ when training \methodname and all NVS counterparts.
The pseudo-images are generated in the same resolution.
All experiments are conducted on NVIDIA L20 GPUs.

\begin{figure*}[t]
  \includegraphics[width=\textwidth]{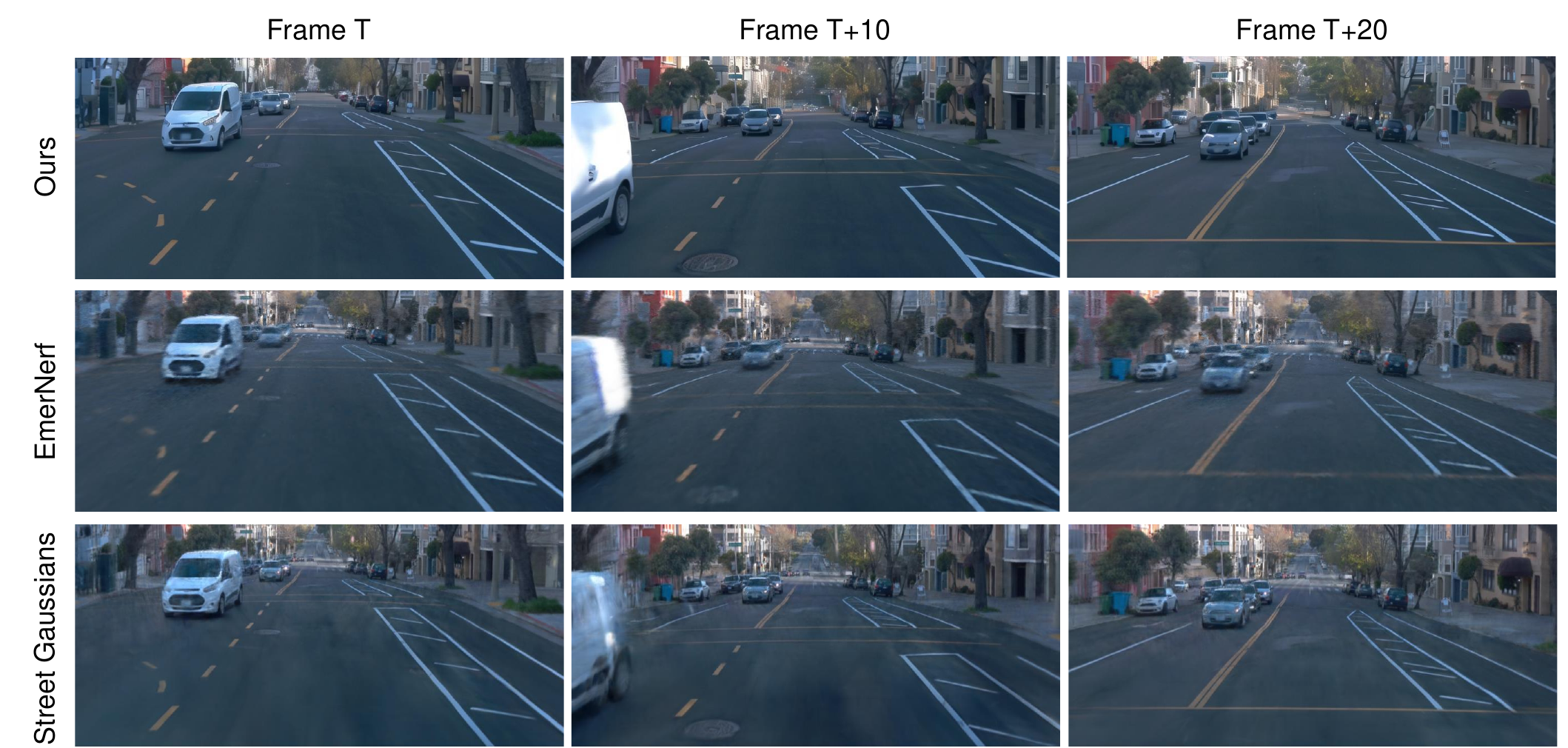}
  \caption{
  \textbf{Visualization comparison on moving objects.} We compare the performance of NVS methods where moving objects are visible in camera views. In the shown case, our proposed method can generate more accurate images of vehicles driving in the opposite lane. Images are synthesized on training views.
  }
  \label{fig:visualize_movingobj}
\end{figure*}
\begin{figure*}[t]
  \includegraphics[width=\textwidth]{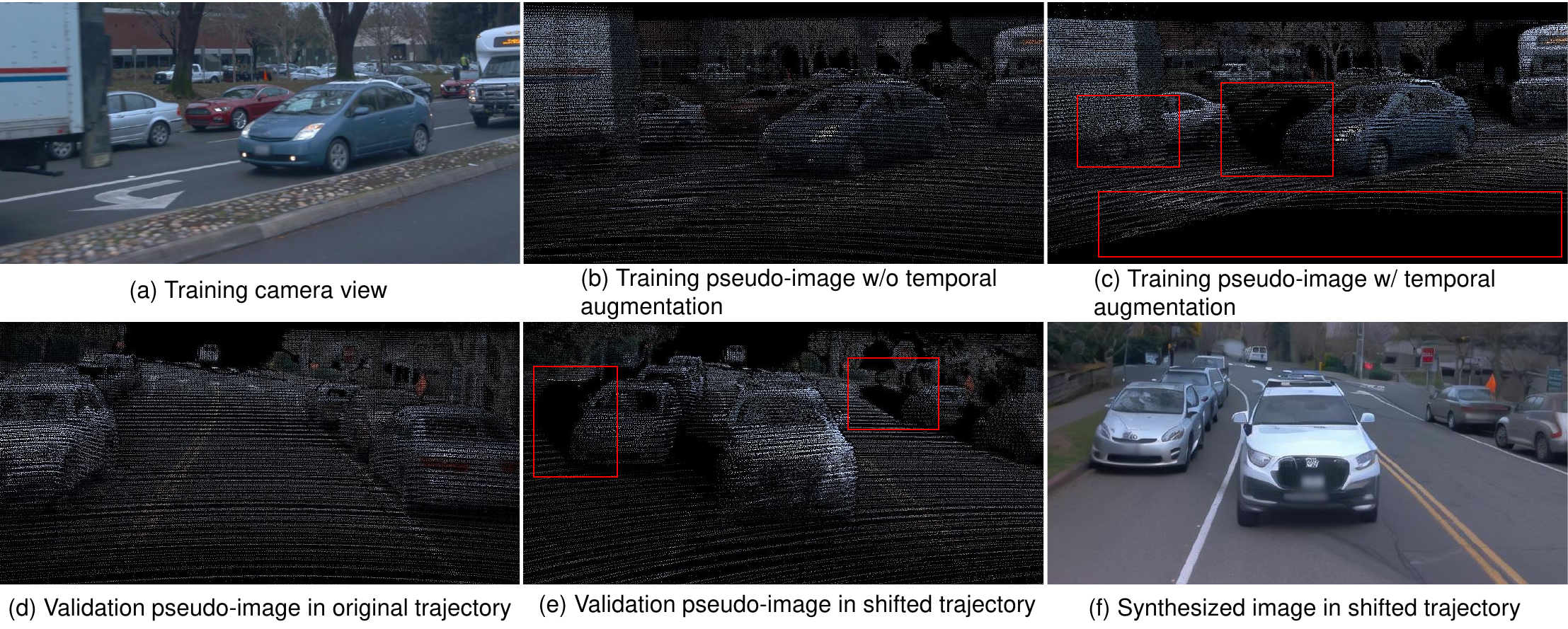}
  \caption{
  \textbf{Visualization of the viewpoint transformation simulation.} For a training sample, we show the (a) source camera view, (b) pseudo-image generated from current LiDAR observation, and (c) pseudo-image generated from previous LiDAR observation to simulate viewpoint transformation.
  We also show a validation case of (d) pseudo-image on the original viewpoint, (e) pseudo-image on the shifted viewpoint, and (f) generated image on the shifted viewpoint.
  Note the image areas with missing or overlapped 3D observation circled by red boxes in (c) and (f).
  The proposed viewpoint transformation simulation on pseudo-images can well-stimulate the insufficient 3D prior observations brought by the shift in viewpoint.
  }
  \label{fig:visualize_temporalaug}
\end{figure*}
\subsection{Video comparison.}
We present a video comparison of novel view sequences synthesized by NVS methods on the modified trajectory in a driving sequence.
Please check the video file submitted as supplementary material.

\subsection{Visualiation comparison on dynamic objects.}
Without the scene reconstruction process, \methodname is free from complex cross-frame optimization of dynamic objects.
\methodname can synthesize images of dynamic objects in high accuracy, as shown in Fig.\ref{fig:visualize_movingobj}.
In comparison, despite with specific design, current reconstruction-based NVS methods still suffer from dynamic object modeling.

\subsection{Visualization of the viewpoint transformation simulation.}
Besides quantitatively ablating the impact of the proposed viewpoint transformation simulation with pseudo-images, we also provide a visualization case in Fig.\ref{fig:visualize_temporalaug} to illustrate its effect.

When facing the translation of viewpoints, pseudo images generated from the existing viewpoints cannot provide complete 3D priors, such as in areas circled by red boxes in Fig.\ref{fig:visualize_temporalaug}(e). To strengthen the robustness of \methodname towards those patterns in pseudo-images, we propose to employ viewpoint transformation simulation with pseudo-images by generating pseudo-images in training views from mismatched LiDAR observations. As shown in Fig.\ref{fig:visualize_temporalaug}(c), we can stimulate the pseudo-image areas with insufficient 3D priors. The generative model trained with the proposed viewpoint transformation simulation with pseudo-images can render images of high quality when facing insufficient 3D prior inputs, as shown in Fig.\ref{fig:visualize_temporalaug}(f).

\subsection{Effect of classifier-free guidance.} 
\begin{figure*}[t]
  \includegraphics[width=\textwidth]{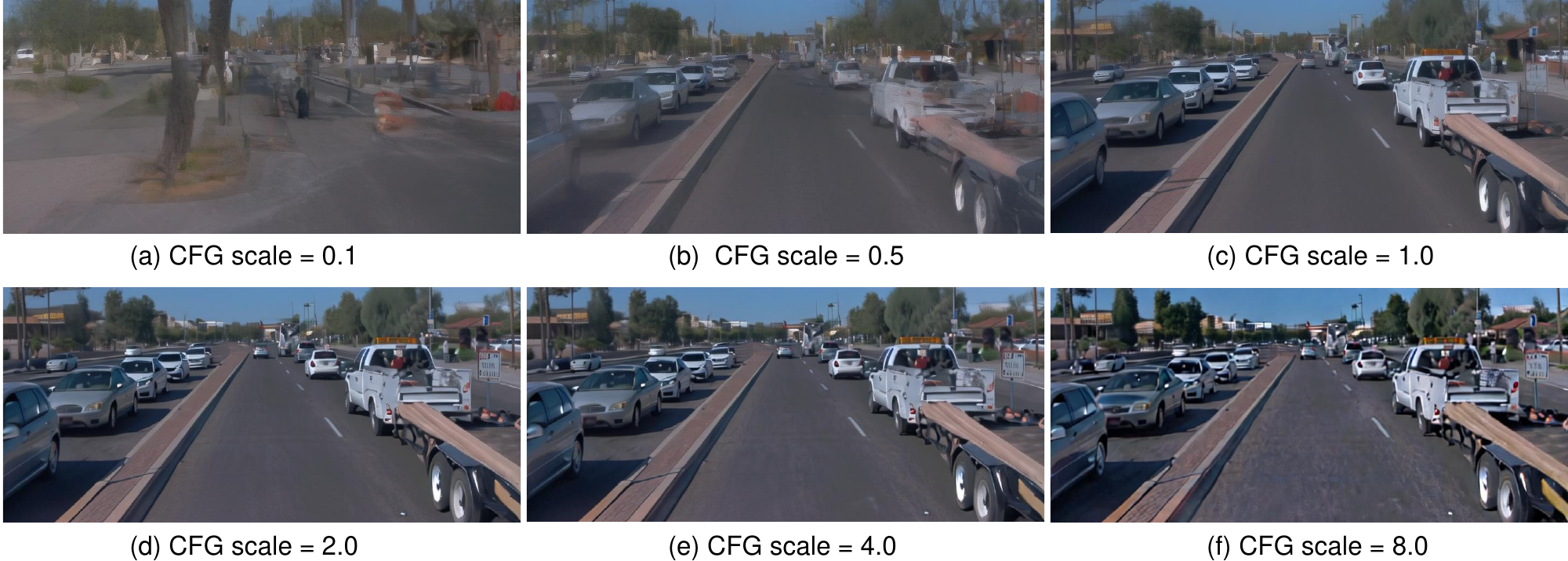}
  \caption{
  \textbf{Visualization of the effect of classifier-free guidance (CFG).} We show generation results with the same input pseudo-image and different CFG scales. The larger CFG scale is set, the stronger the generation result is constrained by the 3D prior condition.
  }
  \label{fig:visualize_cfgabl}
\end{figure*}
As a diffusion model, \methodname can use the classifier-free guidance (CFG) technique to adjust the control effect of input 3D prior condtions.
We show the impact of CFG with different CFG scales in Fig.~\ref{fig:visualize_cfgabl}. 

\begin{figure*}[t]
  \includegraphics[width=\textwidth]{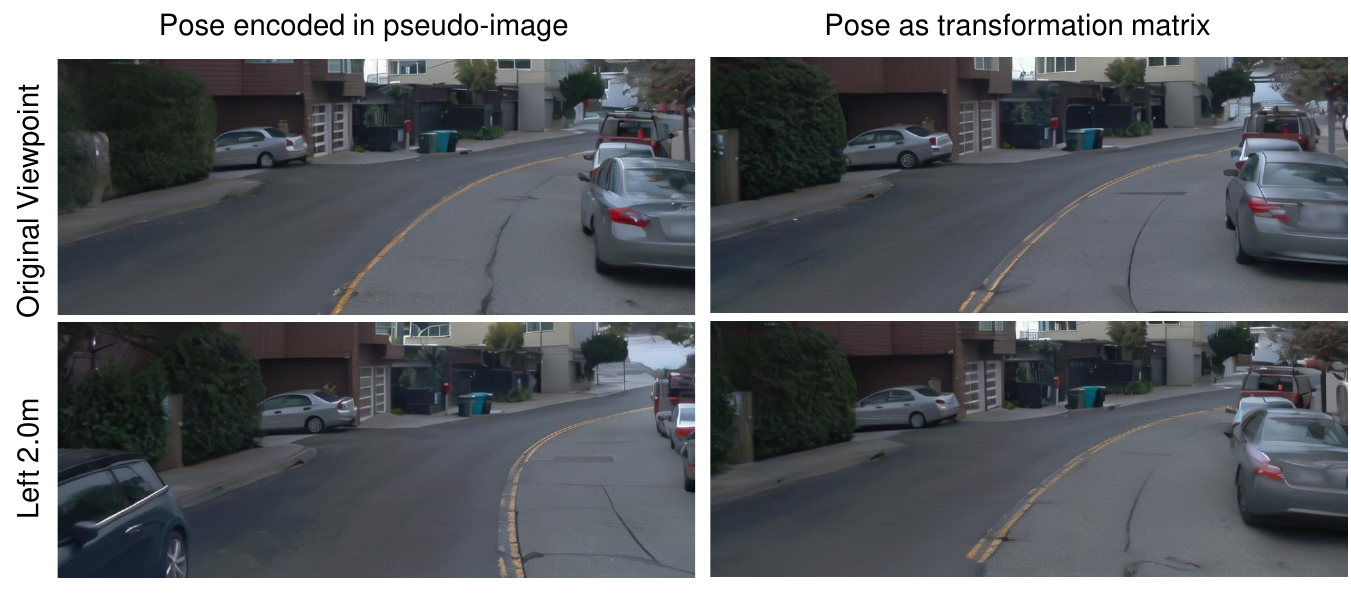}
  \caption{
  \textbf{Visualization comparison on the encoding of camera viewpoint condition.} We qualitatively compare the image synthesis performance of the diffusion model with pseudo-image as input (Table \ref{tab:abl_3dpripr}(a)), or with the reference image, LiDAR points, and viewpoint transformation matrix as input (Table \ref{tab:abl_3dpripr}(d)).
  When feeding the model with pose transformation matrices, the diffusion models often fail to generate views on the targeted viewpoint, as shown in the second column.
  }
  \label{fig:visualize_ablexp}
\end{figure*}
\subsection{Visualization results of component ablation experiment on pseudo-image representation.} For the ablation setting (a) and (d) introduced in Sec.\ref{4.3}, whose quantitative results are reported in Table \ref{tab:abl_3dpripr}, we present a visualization comparison in Fig.\ref{fig:visualize_ablexp}.
Setting (a) is the baseline setting of feeding diffusion models with pseudo-image scene representations for view synthesis.
Under setting (d), we feed the diffusion model with the reference image, LiDAR point cloud, and the transformation matrix from the reference view to the target view. 
As shown in Fig.\ref{fig:visualize_ablexp}, model feed with pseudo-image can precisely synthesize image on the target viewpoint, while model feed with raw camera pose fails to follow the viewpoint condition. 
Given that the diffusion model can only be trained on recorded trajectories, we found the diffusion model fed with reference images and viewpoint pose tends to overfit to the specific camera movement pattern in each camera position.
As shown in Fig.\ref{fig:visualize_ablexp}, the model fed with raw 3D prior inputs will only move the frontal camera view forward or backward, ignoring the viewpoint pose condition.
This is due to the absence of a training sample where the frontal camera is moved laterally.
By modifying the novel view synthesis task as an image completion task based on the pseudo-image representation of 3D priors as well as applying the proposed viewpoint transformation simulation with pseudo-images, we can overcome this training data shortage.

\subsection{Validation Sequences}
We list the selected 12 validation sequences from the WOD dataset here with their official individual file names:
\begin{itemize}
    \item segment-10588771936253546636\_2300\_000\_2320\_000\_with\_camera\_labels.tfrecord,
        \item segment-6242822583398487496\_73\_000\_93\_000\_with\_camera\_labels.tfrecord,
        \item segment-16801666784196221098\_2480\_000\_2500\_000\_with\_camera\_labels.tfrecord,
        \item segment-1191788760630624072\_3880\_000\_3900\_000\_with\_camera\_labels.tfrecord,
        \item segment-10625026498155904401\_200\_000\_220\_000\_with\_camera\_labels.tfrecord,
        \item segment-11846396154240966170\_3540\_000\_3560\_000\_with\_camera\_labels.tfrecord,
        \item segment-18111897798871103675\_320\_000\_340\_000\_with\_camera\_labels.tfrecord,
        \item segment-11017034898130016754\_697\_830\_717\_830\_with\_camera\_labels.tfrecord,
        \item segment-10963653239323173269\_1924\_000\_1944\_000\_with\_camera\_labels.tfrecord,
        \item segment-12161824480686739258\_1813\_380\_1833\_380\_with\_camera\_labels.tfrecord,
        \item segment-11928449532664718059\_1200\_000\_1220\_000\_with\_camera\_labels.tfrecord,
        \item segment-10444454289801298640\_4360\_000\_4380\_000\_with\_camera\_labels.tfrecord.
\end{itemize}

\end{document}